%% file: egbib.tex
\documentclass[10pt,twocolumn,letterpaper]{article}

\usepackage{cvpr}
\usepackage{times}
\usepackage{epsfig}
\usepackage{graphicx}
\usepackage{amsmath}
\usepackage{amssymb}
\usepackage{multirow}
\usepackage{bbm}
\usepackage{makecell}
\newcommand{\PreserveBackslash}[1]{\let\temp=\\#1\let\\=\temp}
\newcolumntype{C}[1]{>{\PreserveBackslash\centering}p{#1}}
\newcommand{\trans}[1]{{#1}^{\ensuremath{\mathsf{T}}}} % transpose
\usepackage[pagebackref=true,breaklinks=true,letterpaper=true,colorlinks,bookmarks=false]{hyperref}

\cvprfinalcopy % *** Uncomment this line for the final submission

\def\cvprPaperID{901} % *** Enter the CVPR Paper ID here

\newcommand\blfootnote[1]{%
  \begingroup
  \renewcommand\thefootnote{}\footnote{#1}%
  \addtocounter{footnote}{-1}%
  \endgroup
}

\ifcvprfinal\fi
\begin{document}

%%%%%%%%% TITLE
\title{Inter-Region Affinity Distillation for Road Marking Segmentation}

\author{\textbf{Yuenan Hou}$^{1}$, \textbf{Zheng Ma}$^{2}$, \textbf{Chunxiao Liu}$^{2}$, \textbf{Tak-Wai Hui}$^{1}$, \textbf{and Chen Change Loy}$^{3\dagger}$\\
$^{1}$The Chinese University of Hong Kong $^{2}$SenseTime Group Limited $^{3}$Nanyang Technological University\\
$^{1}$\{hy117, twhui\}@ie.cuhk.edu.hk, $^{2}$\{mazheng, liuchunxiao\}@sensetime.com, $^{3}$ccloy@ntu.edu.sg
}

\maketitle
\def\algorithmname{IntRA-KD}

%%%%%%%%% ABSTRACT
\begin{abstract}
\input{abstract.tex}
\end{abstract}

%////////////////////////////////
\section{Introduction}
\label{sec:introduction}
%////////////////////////////////
\input{introduction.tex}

%////////////////////////////////
\section{Related Work}
\label{sec:relatedwork}
%////////////////////////////////
\input{relatedwork.tex}

%////////////////////////////////
\section{Methodology}
\label{sec:methodology}
%////////////////////////////////
\input{methodology.tex}

%////////////////////////////////
\section{Experiments}
\label{sec:experiments}
%////////////////////////////////
\input{experiment.tex}

\section{Conclusion}\label{conclusion}

We have proposed a simple yet effective distillation approach, \ie, \algorithmname~, to effectively transfer scene structural knowledge from a teacher model to a student model. The structural knowledge is represented as an inter-region affinity graph to capture similarity of feature distribution of different scene regions. We applied \algorithmname~to various lightweight models and observed consistent performance gains to these models over other contemporary distillation methods. Promising results on three large-scale road marking segmentation benchmarks strongly suggest the effectiveness of \algorithmname. Results on Cityscapes are provided in the supplementary material. %It would be interesting to extend this idea to other tasks that demands inter-region relationship, such as image saliency detection and image matting.

\noindent \textbf{Acknowledgement:} This work is supported by the SenseTime-NTU Collaboration Project, Collaborative Research grant from SenseTime Group (CUHK Agreement No. TS1610626 $\&$ No. TS1712093), Singapore MOE AcRF Tier 1 (2018-T1-002-056), NTU SUG, and NTU NAP.

{\small
\bibliographystyle{ieee_fullname}
\bibliography{egbib}
}

\end{document}

%% file: abstract.tex
% !TEX root = ../egpaper_for_review.tex

We study the problem of distilling knowledge from a large deep teacher network to a much smaller student network for the task of road marking segmentation. 
In this work, we explore a novel knowledge distillation (KD) approach that can transfer `knowledge' on scene structure more effectively from a teacher to a student model. Our method is known as Inter-Region Affinity KD (IntRA-KD). It decomposes a given road scene image into different regions and represents each region as a node in a graph. An inter-region affinity graph is then formed by establishing pairwise relationships between nodes based on their similarity in feature distribution.
To learn structural knowledge from the teacher network, the student is required to match the graph generated by the teacher.
The proposed method shows promising results on three large-scale road marking segmentation benchmarks, i.e., ApolloScape, CULane and LLAMAS, by taking various lightweight models as students and ResNet-101 as the teacher.
IntRA-KD consistently brings higher performance gains on all lightweight models, compared to previous distillation methods. Our code is available at \url{https://github.com/cardwing/Codes-for-IntRA-KD}.% such as that based on softened class scores or spatial attention transfer. 

%% file: introduction.tex
% !TEX root = ../egpaper_for_review.tex

\blfootnote{$\dagger$: Corresponding author.}

Road marking segmentation serves various purposes in autonomous driving, \eg, providing cues for vehicle navigation or extracting basic road elements and lanes for constructing high-definition maps~\cite{Homayounfar_2019_ICCV}. 
Training a deep network for road marking segmentation is known to be challenging due to various reasons~\cite{hou2019learning}, including tiny road elements, poor lighting conditions and occlusions caused by vehicles.
The training difficulty is further compounded by the nature of segmentation labels available for training, which are usually sparse (\eg, very thin and long lane marking against a large background), hence affecting the capability of a network in learning the spatial structure of a road scene~\cite{hou2019learning,pan2017spatial}.

The aforementioned challenges become especially crippling when one is required to train a small model for road marking segmentation. This requirement is not uncommon considering that small models are usually deployed on vehicles with limited computational resources.
Knowledge distillation (KD)~\cite{hinton2015distilling} offers an appealing way to facilitate the training of a small student model by transferring knowledge from a trained teacher model of larger capacity.
Various KD methods have been proposed in the past, \eg, with knowledge transferred through softened class scores~\cite{hinton2015distilling}, feature maps matching~\cite{hou2019learningto,liu2019structured} or spatial attention maps matching~\cite{zagoruyko2016paying}.

While existing KD methods are shown effective in many classification tasks, we found that they still fall short in transferring knowledge of scene structure for the task of road marking segmentation. 
Specifically, a road scene typically exhibits consistent configuration, \ie, road elements are orderly distributed in a scene. The structural relationship is crucial to providing the necessary constraint or regularization, especially for small networks, to combat against the sparsity of supervision. However, such structural relationship is rarely exploited in previous distillation methods.
The lack of structural awareness makes small models struggle in differentiating visually similar but functionally different road markings.

In this paper, we wish to enhance the structural awareness of a student model by exploring a more effective way to transfer the scene structure prior encoded in a teacher to a student.
Our investigation is based on the premise that a teacher model should have a better capability in learning discriminative features and capturing contextual information due to its larger capacity in comparison to the student model. Feature distribution relationships encoded by the teacher on different parts of a deep feature map could reveal rich structural connections between different scene regions, \eg, the lane region should look different from the zebra crossing. Such priors can offer a strong constraint to regularize the learning of the student network.

\begin{figure}[t]
 \centering
 \includegraphics[width=1.0\linewidth]{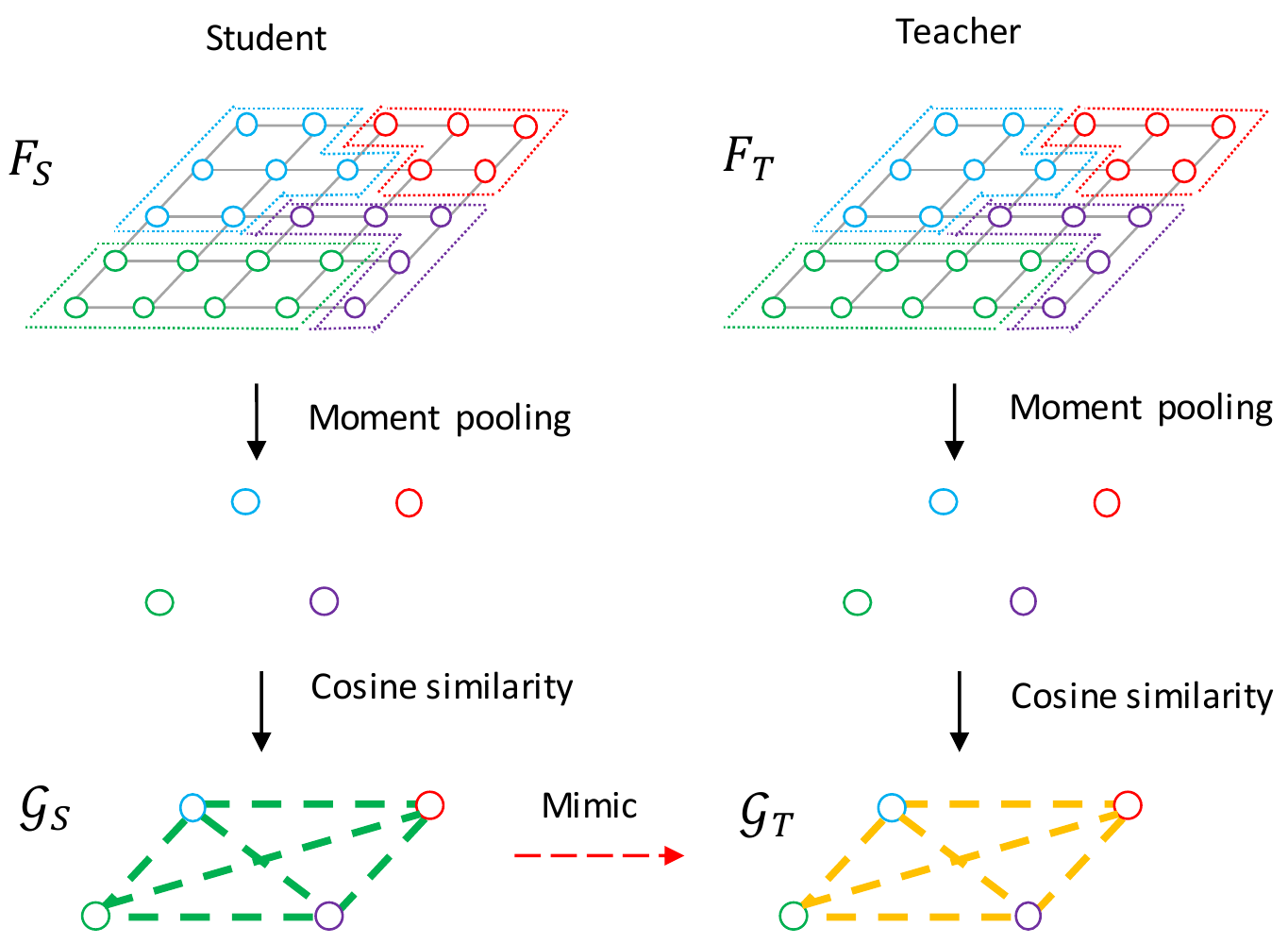}
 \vskip -0.2cm
 \caption{Illustration of the affinity distillation process. $F_{S}$ and $F_{T}$ are the intermediate activations of the student and teacher models, respectively. $\mathcal{G}$ is the affinity graph that comprises nodes (feature vectors) and edges (cosine similarities). Note that each circle in the figure is a vector and different colors represent different classes.}
 \centering
 \vskip -0.4cm
 \label{fig:graph_map}
\end{figure}

Our method is known as \textit{Inter-Region Affinity Knowledge Distillation} (IntRA-KD). 
As the name implies, knowledge on scene structure is represented as inter-region affinity graphs, as shown as Fig.~\ref{fig:graph_map}. Each region is a part of a deep feature map, while each node in the graph denotes the feature distribution statistics of each region. Each pair of nodes are connected by an edge representing their similarity in terms of feature distribution.
Given the same input image, the student network and the teacher network will both produce their corresponding affinity graph. Through graph matching, a distillation loss on graph consistency is generated to update the student network.

This novel notion of inter-region affinity knowledge distillation is appealing in its simplicity and generality. The method is applicable to various road marking scenarios with an arbitrary number of road element classes. It can also work together with other knowledge distillation methods. It can even be applied on more general segmentation tasks (e.g., Cityscapes~\cite{cordts2016cityscapes}).
We present an effective and efficient way of building inter-region affinity graphs, including a method to obtain regions from deep feature maps and a new moment pooling operator to derive feature distribution statistics from these regions.
Extensive experiments on three popular datasets (ApolloScape~\cite{huang2018apolloscape}, CULane~\cite{pan2017spatial} and LLAMAS~\cite{llamas2019}) show that IntRA-KD consistently outperforms other KD methods, \eg, probability map distillation~\cite{hinton2015distilling} and attention map distillation~\cite{zagoruyko2016paying}. It generalizes well to various student architectures, \eg, ERFNet~\cite{romera2017erfnet}, ENet~\cite{paszke2016enet} and ResNet-18~\cite{he2016deep}. Notably, with IntRA-KD, ERFNet achieves compelling performance in all benchmarks with \textbf{21}$\times$ fewer parameters (2.49 M v.s. 52.53 M) and runs \textbf{16}$\times$ faster (10.2 ms v.s. 171.2 ms) compared to ResNet-101 model. Encouraging results are also observed on Cityscapes~\cite{cordts2016cityscapes}. Due to space limit, we include the results in the supplementary material.

%% file: relatedwork.tex
% !TEX root = ../egpaper_for_review.tex

\noindent \textbf{Road marking segmentation.} Road marking segmentation is conventionally handled using hand-crafted features to obtain road marking segments. Then, a classification network is employed to classify the category of each segment~\cite{huang2014lane, qin2013general}. These approaches have many drawbacks, \eg, require sophisticated feature engineering process and only work well in simple highway scenarios. 

The emergence of deep learning has avoided manual feature design through learning features in an end-to-end manner. These approaches usually adopt the dense prediction formulation, \ie, assign each pixel a category label~\cite{hou2019learning, pan2017spatial, wang2019apolloscape}. For example, Wang \etal ~\cite{wang2019apolloscape} exploit deep neural networks to map an input image to a segmentation map. Since large models usually demand huge memory storage and have slow inference speed, many lightweight models, \eg, ERFNet~\cite{romera2017erfnet}, are leveraged to fulfil the requirement of fast inference and small storage~\cite{hou2019learning}. However, due to the limited model size, these small networks perform poorly in road marking segmentation. A common observation is that such small models do not have enough capacity to capture sufficient contextual knowledge given the sparse supervision signals~\cite{hou2019learning,pan2017spatial,zhang2018geometric}. Several schemes have been proposed to relieve the sparsity problem. For instance, Hou \etal ~\cite{hou2019learning} reinforce the learning of contextual knowledge through self knowledge distillation, \ie, using deep-layer attention maps to guide the learning of shallower layers. SCNN~\cite{pan2017spatial} resolves this problem through message passing between deep feature layers. Zhang \etal ~\cite{zhang2018geometric} propose a framework to perform lane area segmentation and lane boundary detection simultaneously. 
The aforementioned methods do not take structural relationship between different areas into account and they do not consider knowledge distillation from teacher networks. %On the contrary, IntTRA-KD establishes the inter-region affinity graph and can effectively utilize the structural information to enhance the performance of small models on road marking segmentation.

\noindent \textbf{Knowledge distillation.} Knowledge distillation was originally introduced by~\cite{hinton2015distilling} to transfer knowledge from a teacher model to a compact student model. The distilled knowledge can be in diverse forms, \eg, softened output logits~\cite{hinton2015distilling}, intermediate feature maps~\cite{gao2018embarrassingly, hou2019learningto, liu2019structured, zhu2018bidirectional} or pairwise similarity maps between neighbouring layers~\cite{yim2017gift}. There is another line of work~\cite{hou2019learning, sun2019deeply} that uses self-derived knowledge to reinforce the representation learning of the network itself, without the supervision of a large teacher model. Recent studies have expanded knowledge distillation from one sample to several samples~\cite{liu2019knowledge, park2019relational, peng2019correlation, tung2019similarity}. For instance, Park \etal ~\cite{park2019relational} transfer mutual relations between a batch of data samples in the distillation process. Tung \etal ~\cite{tung2019similarity} take the similarity scores of features of different samples as distillation targets. The aforementioned approaches~\cite{liu2019knowledge, park2019relational, peng2019correlation, tung2019similarity} do not consider the structural relationship between different areas in one sample. On the contrary, the proposed IntRA-KD takes inter-region relationship into account, which is new in knowledge distillation.

%% file: methodology.tex
% !TEX root = ../egpaper_for_review.tex

\begin{figure*}[t]
  \centering
  \includegraphics[width=\linewidth]{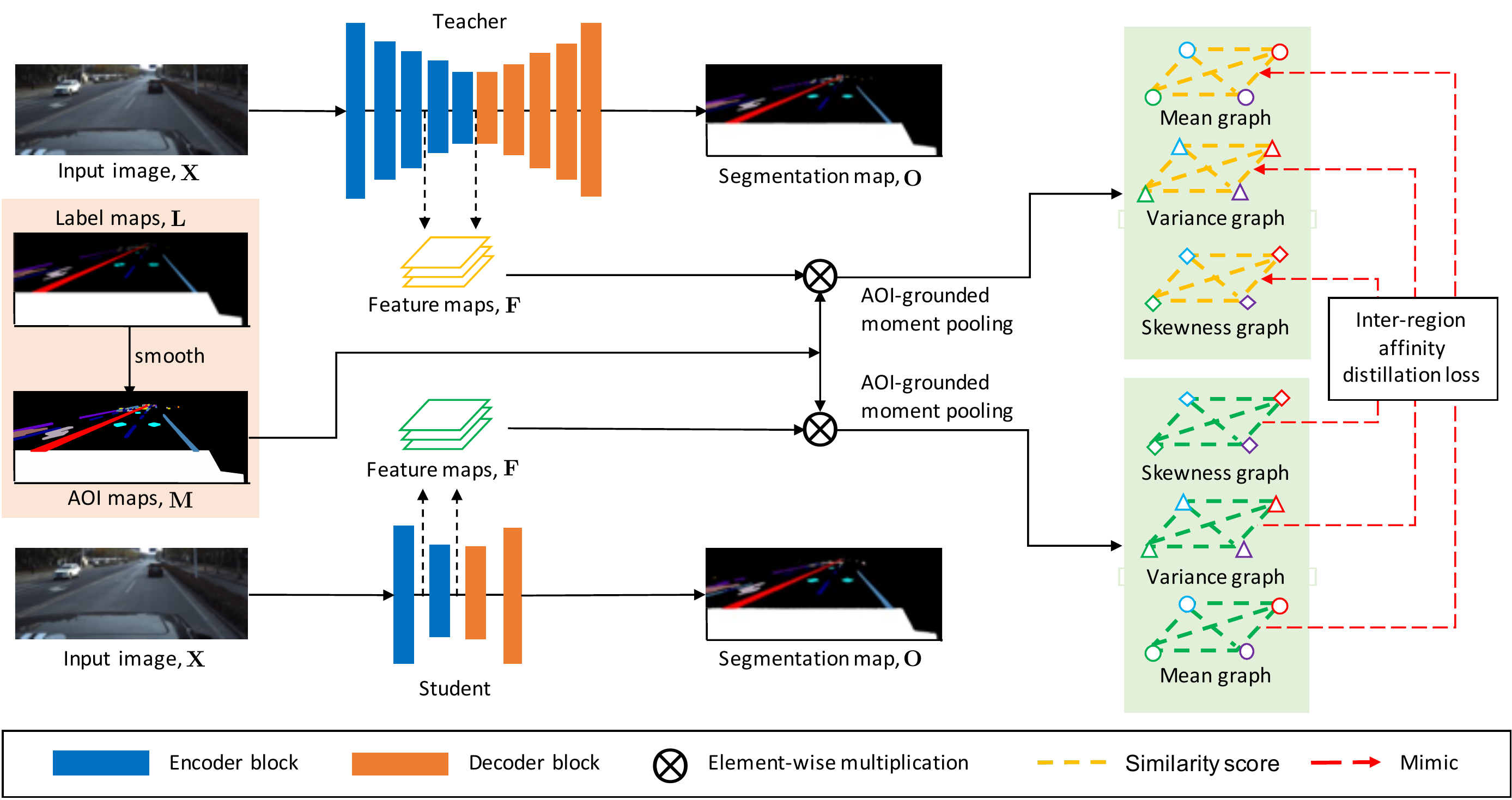}
  \caption{The pipeline of \algorithmname. There are two networks in our approach, one serves as the student and the other serves as the teacher. Given an input image, the student is required to mimic the inter-region affinity graph of the trained teacher model at selected layers. %The student can optionally mimic the attention maps of the teacher too, which can be derived following~\cite{zagoruyko2016paying}. 
  Labels are pre-processed by a smoothing operation to obtain the areas of interest (AOI). AOI maps, shown as an integrated map here, provide the masks to extract features corresponding to each class region. Moment pooling is performed to compute the statistics of feature distribution for each region. This is followed by the construction of an inter-region affinity graph that captures the similarity of feature distribution between different regions. The inter-region affinity graph is composed of three sub-graphs, \ie, the mean graph, the variance graph and the skewness graph.}
  \centering 
  \vskip -0.4cm
  \label{fig:pipeline}
\end{figure*}

Road marking segmentation is commonly formulated as a semantic segmentation task~\cite{wang2019apolloscape}. More specifically, given an input image $\textbf{X} \in \mathbb{R}^{h \times w \times 3}$, the objective is to assign a label $l \in \{0, \dots, n-1\}$ to each pixel $(i, j)$ of $\textbf{X}$, comprising the segmentation map $\textbf{O}$. Here, $h$ and $w$ are the height and width of the input image, $n$ is the number of classes and class 0 denotes the background. The objective is to learn a mapping $\mathcal{F}$: $\textbf{X} \mapsto \textbf{O}$. Contemporary algorithms use CNN as $\mathcal{F}$ for end-to-end prediction. 

Since autonomous vehicles have limited computational resources and demand real-time performance, lightweight models are adopted to fulfil the aforementioned requirements. On account of limited parameter size as well as insufficient guidance due to sparse supervision signals, these small models usually fail in the challenging road marking segmentation task. Knowledge distillation~\cite{hinton2015distilling, hou2019learning, liu2019structured} is a common approach to improving the performance of small models by means of distilling knowledge from large models.  There are two networks in knowledge distillation, one is called the student and the other is called the teacher. The purpose of knowledge distillation is to transfer dark knowledge from the large, cumbersome teacher model to the small, compact student model. The dark knowledge can take on many forms, \eg, output logits and intermediate layer activations. 
There exist previous distillation methods~\cite{park2019relational, peng2019correlation, tung2019similarity} that exploit the relationship between a batch of samples. These approaches, however, do not take into account the structural relationship between different areas within a sample.

\subsection{Problem Formulation}
 
Unlike existing KD approaches, \algorithmname~considers intrinsic structural knowledge within each sample as a form of knowledge for distillation.
Specifically, we consider each input sample to have $n$ road marking classes including the background class. We treat each class map as a region. In practice, the number of classes/regions co-existing in a sample can be fewer than $n$.
Given the same input, an inter-region affinity graph $\mathcal{G}_S$ for the student network and an inter-region affinity graph $\mathcal{G}_T$ for the teacher are constructed. Here, an affinity graph is defined as 
\begin{equation}
\label{eqn:graph}
\mathcal{G} = \langle \boldsymbol{\mu}, \mathbf{C} \rangle,
\end{equation}
where $\boldsymbol{\mu}$ is a set of nodes, each of which represents feature distribution statistics of each region. Each pair of nodes are connected by an edge $\mathbf{C}$ that denotes the similarity between two nodes in terms of feature distribution.

The overall pipeline of our \algorithmname~is shown in Fig.~\ref{fig:pipeline}.
The framework is composed of three main components: 

\noindent
1) \textit{Generation of areas of interest (AOI)} -- to extract regions representing the spatial extent for each node in the graphs.

\noindent
2) \textit{AOI-grounded moment pooling} -- to quantify the statistics of feature distribution of each region. 

\noindent
3) \textit{Inter-region affinity distillation} -- to construct the inter-region affinity graph and distill structural knowledge from the teacher to the student.

\subsection{Inter-Region Affinity Knowledge Distillation}
%We use the subscript to differentiate the output of the teacher model from that of the student model.
%Unlike conventional semantic segmentation~\cite{Long_2015_CVPR, Sun_2019_CVPR} that performs segmentation on the whole image, road markings only account for a small part of a scene. 
%
%Hence, we need to generate AOI and perform distillation in these areas, which is more effective.

\begin{figure}[t]
  \centering
  \includegraphics[width=1.0\linewidth]{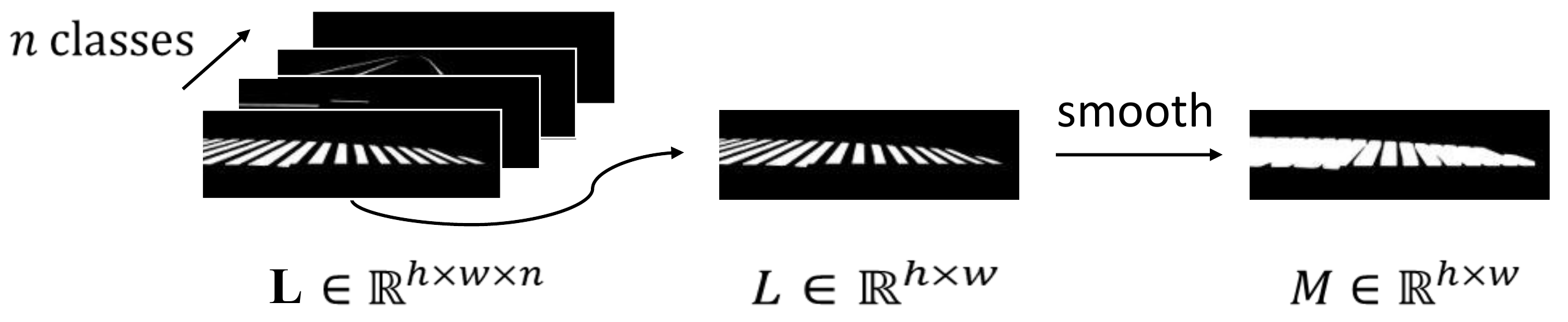}
  \caption{Generation of AOI. Take one class label map $L$ as example. We perform smoothing on $L$ and obtain one AOI map $M$.}
  \centering
  \vskip -0.4cm
  \label{fig:aoi}
\end{figure}

\noindent \textbf{Generation of AOI.} The first step of IntRA-KD is to extract regions from a given image to represent the spatial extent of each class. The output of this step is $n$ AOI maps constituting a set $\mathbf{M} \in \mathbb{R}^{h \times w \times n}$, where $h$ is the height, $w$ is the width, and $n$ is the number of classes.
Each mask map is binary -- `1' represents the spatial extent of a particular class, \eg, left lane, while `0' represents other classes and background.

A straightforward solution is to use the ground-truth labels as AOI. However, ground-truth labels only consider the road markings but neglect the surrounding areas around the road markings. 
We empirically found that na\"{i}ve distillation in ground-truth areas is ineffective for the transfer of contextual information from a teacher to a student model. 

To include a larger area, we use a transformation operation to generate AOI from the ground-truth labels. Unlike labels that only contain road markings, areas obtained after the operation also take into account the surrounding areas of road markings.
An illustration of AOI generation is shown in Fig.~\ref{fig:aoi}.
Suppose we have $n$ binary ground-truth label maps comprising a set $\mathbf{L} \in \mathbb{R}^{h \times w \times n}$. 
For each class label map $L \in \mathbb{R}^{h \times w}$, we smooth the label map with an average kernel $\phi$ and obtain an AOI map for the corresponding class as $M = \mathbbm{1}\left(\phi(L)>0\right)$, where $\mathbbm{1}(.)$ is an indicator function and $M \in \mathbb{R}^{h \times w}$ has the same size as $L$. 
Repeating these steps for all $n$ ground-truth label maps provides us $n$ AOI maps.
Note that AOI maps can also be obtained by image morphological operations.

\vspace{0.1cm}
\noindent \textbf{AOI-grounded moment pooling.} 
Suppose the feature maps of a network are represented as $\mathbf{F} \in \mathbb{R}^{h_{f} \times w_{f} \times c}$, where $h_{f}$, $w_{f}$ and $c$ denote the height, width and channel of the feature map, respectively. 
Once we obtain the AOI maps $\mathbf{M}$, we can use them as masks to extract AOI features from $\mathbf{F}$ for each class region. 
The obtained AOI features can then be used to compute the inter-region affinity.
For effective affinity computation, we regard AOI features of each region as a distribution. Affinity can then be defined as the similarity between two feature distributions.

Moments have been widely-used in many studies~\cite{Peng_2019_ICCV,zellinger2017central}. 
Inspired by these prior studies, we calculate moment statistics of a distribution and use them for affinity computation. In particular, we extract the first moment $\boldsymbol{\mu}_{1}$, second moment $\boldsymbol{\mu}_{2}$ and third moment $\boldsymbol{\mu}_{3}$ as the high-level statistics of a distribution. 
The moments of features have explicit meanings, \ie, the first moment represents the mean of the distribution, the second moment (variance) and the third moment (skewness) describe the shape of that distribution.
We empirically found that using higher-order moments brings marginal performance gains while requiring heavier computation cost.

To compute $\boldsymbol{\mu}_{1}(k)$, $\boldsymbol{\mu}_{2}(k)$ and $\boldsymbol{\mu}_{3}(k)$ of class $k$, we introduce the \textit{moment pooling operation} to process the AOI features. 
%
%\begin{equation}
\begin{align}
\label{eqn:pooling_op}
\begin{split}
\boldsymbol{\mu}_{1}(k) &= \frac{1}{|\textbf{M}(:, :, k)|} \sum_{i=1}^{h_{f}} \sum_{j=1}^{w_{f}} \mathbf{M}(i, j, k) \mathbf{F}(i, j),
\\
\boldsymbol{\mu}_{2}(k) &= \frac{1}{|\textbf{M}(:, :, k)|} \sum_{i=1}^{h_{f}} \sum_{j=1}^{w_{f}} (\mathbf{M}(i, j, k) \mathbf{F}(i, j) - \boldsymbol{\mu}_{1}(k))^{2},
\\
\boldsymbol{\mu}_{3}(k) &= \frac{1}{|\mathbf{M}(:, :, k)|} \sum_{i=1}^{h_{f}} \sum_{j=1}^{w_{f}} \left(\frac{\mathbf{M}(i, j, k) \mathbf{F}(i, j) - \boldsymbol{\mu}_{1}(k)}{\boldsymbol{\mu}_{2}(k)}\right)^{3},
\end{split}
\end{align}
%\end{equation}
where $|\mathbf{M}(:, :, k)|$ computes the number of non-zero elements in $\mathbf{M}(:, :, k)$ and $\boldsymbol{\mu}_{r}(k) \in \mathbb{R}^{c}, r \in \{1, 2, 3\}$. 

\begin{figure}[t]
  \centering
  \includegraphics[width=0.8\linewidth]{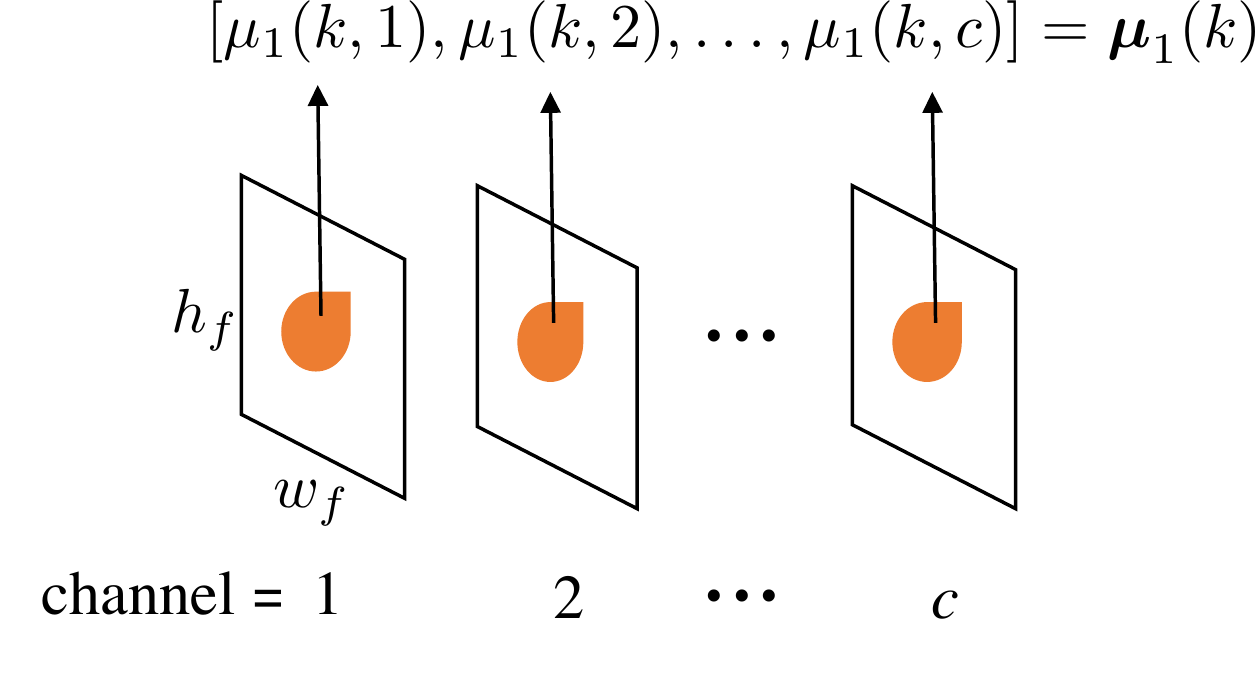}
  \caption{Performing moment pooling on deep feature $\mathbf{F} \in \mathbb{R}^{h_{f} \times w_{f} \times c}$. We use the pooling of first moment as an example.}
  \centering
  \vskip -0.4cm
  \label{fig:moment_pooling}
\end{figure}

An illustration of the process of moment pooling is depicted in Fig.~\ref{fig:moment_pooling}.
The moment pooling operation has the following properties. First, it can process areas with arbitrary shapes and sizes, which can be seen as an extension of the conventional average pooling. Second, the moment vectors obtained by the moment pooling operation can faithfully reflect the feature distribution of a particular region, and yet, the vectors are in a very low-dimension, thus facilitating efficient affinity computation in the subsequent step. 
%Third, 
%%the extracted knowledge is relatively comprehensive since the moment vectors take all pixels in the AOI into consideration. 
%the moment pooling operation can also help neutralize the label noise caused by human annotators since it takes all pixels in the AOI features into account. 

\vspace{0.1cm}
\noindent \textbf{Inter-region affinity distillation.}
Since output feature maps of the teacher model and those of the student model may have different dimensions, performing matching of each pair of moment vectors would require extra parameters or operations to guarantee dimension consistency. Instead, we compute the cosine similarity of the moment vectors of class $k_{1}$ and class $k_{2}$, \ie, 
\begin{equation}
\label{eqn:inter_region_mean_loss}
\mathbf{C}(k_{1}, k_{2}, r) = \frac{\trans{\boldsymbol{\mu}_{r}(k_{1})} \boldsymbol{\mu}_{r}(k_{2})}{\|\boldsymbol{\mu}_{r}(k_{1})\|_{2}\|\boldsymbol{\mu}_{r}(k_{2})\|_{2}}, r \in \{1, 2, 3\}.
\end{equation}

The similarity score captures the similarity of each pair of classes and it is taken as the high-level knowledge to be learned by the student model. The moment vectors $\boldsymbol{\mu}$ and the similarity scores $\mathbf{C}$ constitute the nodes and the edges of the affinity graph $\mathcal{G}= \langle \boldsymbol{\mu}, \mathbf{C} \rangle$, respectively (see Fig.~\ref{fig:pipeline}). 
The inter-region affinity distillation loss is given as follows:
\begin{equation}
\label{eqn:inter_region_mean_loss}
\begin{split}
& \mathcal{L}_{m}(\mathbf{C}_{S}, \mathbf{C}_{T}) = \\
& \frac{1}{3n^{2}} \sum_{r=1}^{3} \sum_{k_{1}=1}^{n} \sum_{k_{2}=1}^{n} \| \mathbf{C}_{S}(k_{1}, k_{2}, r) - \mathbf{C}_{T}(k_{1}, k_{2}, r) \|_{2}^{2}.
\end{split}
\end{equation}
%\vskip -0.4cm

The introduced affinity distillation is robust to the network differences between the teacher and student models since the distillation is only related to the number of classes and is irrelevant to the specific dimension of feature maps. In addition, the affinity knowledge is comprehensive since it gathers information from AOI features from both the foreground and background areas. Finally, in comparison to previous distillation methods~\cite{hinton2015distilling} that use probability maps as distillation targets, the affinity graph is more memory-efficient since it reduces the size of the distillation targets from $h \times w \times n$ to $n^{2}$ where $n$ is usually thousands of times smaller than $h \times w$.

\begin{figure}[t]
  \centering
  \includegraphics[width=1.0\linewidth]{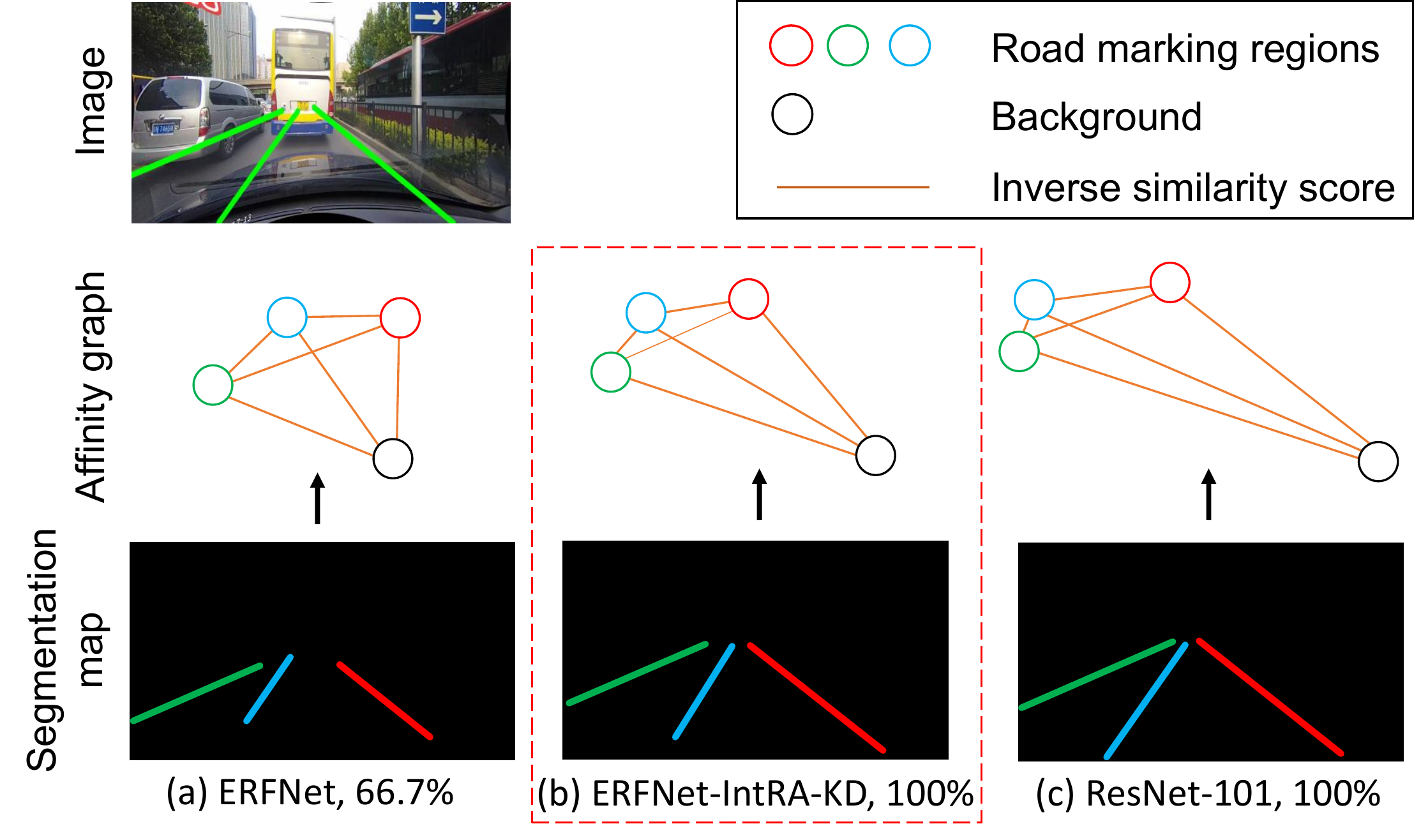}
  \vskip -0.2cm
  \caption{Visualization of the affinity graph generated by different methods. We represent edges in an affinity graph by inverse similarity score. The number next to the method's name is $F_{1}$-measure.}
  \centering
  \vskip -0.6cm
  \label{fig:score}
\end{figure}

From Fig.~\ref{fig:score}, we can see that \algorithmname~not only improves the predictions of ERFNet, but also causes a closer feature structure between the student model and the ResNet-101 teacher model. This is reflected by the very similar structure between the affinity graphs of ERFNet and ResNet-101.
It is interesting to see that those spatially close and visually similar road markings are pulled closer and those spatially distant and visually different markings are pulled apart in the feature space using \algorithmname~. An example is shown in Fig.~\ref{fig:score}, illustrating the effectiveness of \algorithmname~in transferring structural knowledge from the teacher model to the student model. We show in the experiment section that such transfers are essential to improve the performance of student models.

\noindent \textbf{Adding \algorithmname~to training.} The final loss is composed of three terms, \ie, the cross entropy loss, the inter-region affinity distillation loss and the attention map distillation loss. The attention map distillation loss is optional in our framework but it is useful to complement the region-level knowledge. The final loss is written as
\begin{equation}
\label{eqn:total_loss}
\begin{split}
& \mathcal{L} = \mathcal{L}_\mathrm{seg}(\mathbf{O}, \mathbf{L}) + \alpha_{1} \mathcal{L}_\mathrm{m}(\mathbf{C}_{S}, \mathbf{C}_{T}) + \alpha_{2} \mathcal{L}_\mathrm{a}(\mathbf{A}_{S}, \mathbf{A}_{T}).
\end{split}
\end{equation}

\noindent Here, $\alpha_{1}$ and $\alpha_{2}$ are used to balance the effect of different distillation losses on the main task loss $\mathcal{L}_\mathrm{seg}$. % and $\textbf{O}_{l} \in \mathbb{R}^{h \times w}$ is the ground-truth label.
Different from the mimicking of feature maps $\mathbf{F} \in \mathbb{R}^{h_{f} \times w_{f} \times c}$, which demand huge memory resources and are hard to learn, attention maps $\mathbf{A} \in \mathbb{R}^{h_{f} \times w_{f}}$ are more memory-friendly and easier to mimic since only several important areas are needed to learn. The attention map distillation loss is given as follows:
\begin{equation}
\label{eqn:pixel_loss_attention}
\begin{split}
\mathcal{L}_{a}( \mathbf{A}_{S}, \mathbf{A}_{T}) = \sum_{i=1}^{h_{f}} \sum_{j=1}^{w_{f}} \| \mathbf{A}_{S}(i, j) - \mathbf{A}_{T}(i, j) \|_{2}^{2}.
\end{split}
\end{equation}
\noindent We follow~\cite{zagoruyko2016paying} to derive attention maps from feature maps.

%% file: experiment.tex
% !TEX root = ../egpaper_for_review.tex

\begin{table*}[!t]
\caption{Basic information of three road marking segmentation datasets.}
\vskip 0.1cm
\label{dataset_table}
\centering
\small{
\begin{tabular}{c|c|c|c|c|c|c|c}
\hline
Name & \# Frame & Train & Validation & Test & Resolution & \# Class & Temporally continuous ? \\
\hline
\hline
ApolloScape~\cite{huang2018apolloscape} & 114, 538 & 103, 653 & 10, 000 & 885 & 3384 $\times$ 2710 & 36 & $\surd$ \\
%\hline
CULane~\cite{pan2017spatial} & 133, 235 & 88, 880 & 9, 675 & 34, 680 & 1640 $\times$ 590 & 5 & $\times$ \\
%\hline
LLAMAS~\cite{llamas2019} & 79, 113 & 58, 269 & 10, 029 & 10, 815 & 1276 $\times$ 717 & 5 & $\surd$ \\
\hline
\end{tabular}
}
\vspace{-4ex}
\end{table*}

\begin{figure}[t]
  \centering
  \includegraphics[width=1.0\linewidth]{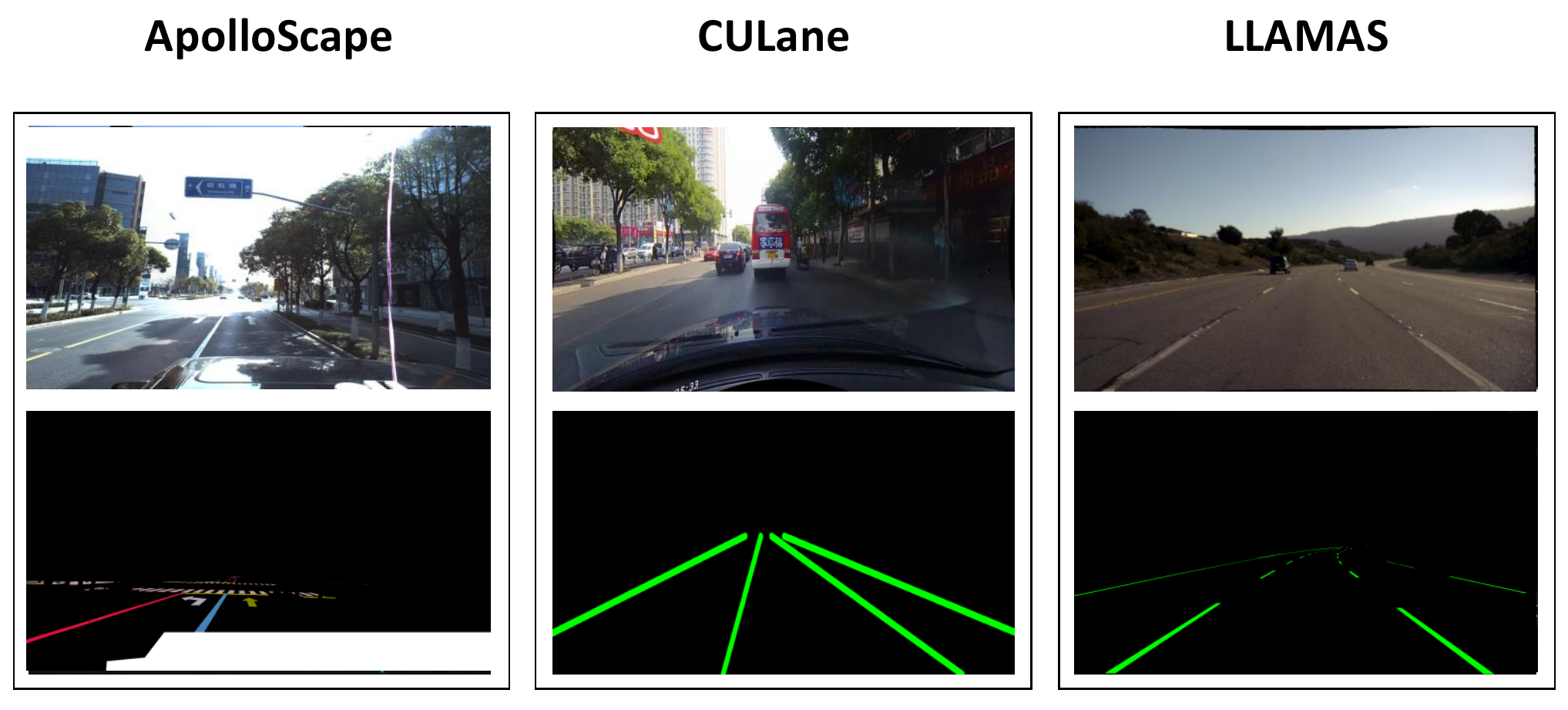}
  \vskip -0.2cm
  \caption{Typical video frames of ApolloScape, CULane and LLAMAS datasets.}
  \centering
  \vskip -0.5cm
  \label{fig:dataset}
\end{figure}

\vspace{0.1cm}
\noindent
\textbf{Datasets}.
We conduct experiments on three datasets, namely ApolloScape~\cite{huang2018apolloscape}, CULane~\cite{pan2017spatial} and LLAMAS~\cite{llamas2019}.
Figure~\ref{fig:dataset} shows a selected video frame from each of the three datasets.
These three datasets are challenging due to poor light conditions, occlusions and the presence of many tiny road markings. 
Note that CULane and LLAMAS only label lanes according to their relative positions to the ego vehicle while ApolloScape labels every road marking on the road according to their functions. Hence, ApolloScape has much more classes and it is more challenging compared with the other two datasets. Apart from the public result~\cite{huang2018apolloscape}, we also reproduce the most related and state-of-the-art methods (\eg, ResNet-50 and UNet-ResNet-34) on ApolloScape for comparison. As to LLAMAS dataset, since the official submission server is not established, the evaluation on the original testing set is impossible. Hence, we split the original validation set into two parts, \ie, one is used for validation and the other is used for testing. Table~\ref{dataset_table} summarizes the details and train/val/test partitions of the datasets.
%
%We will release more details of dataset partitions after peer review.     

\noindent
\textbf{Evaluation metrics}. We use different metrics on each dataset following the guidelines of the benchmark and practices of existing studies.

\noindent \textit{1) ApolloScape.} We use the official metric, \ie, mean intersection-over-union (mIoU) as evaluation criterion~\cite{huang2018apolloscape}.

\noindent \textit{2) CULane.} Following~\cite{pan2017spatial}, we use $F_{1}$-measure as the evaluation metric, which is defined as: $F_{1} = \frac{2 \times Precision \times Recall}{Precision + Recall}$,
%to judge whether a lane is correctly detected, we compute the IoU between ground-truth labels and model predictions. Predictions whose IoUs are larger than 0.5 are considered as true positives (TP). Then, treat each lane as a line with 30 pixel width and
%
%\begin{equation}
%F_{1} = \frac{2 \times Precision \times Recall}{Precision + Recall} ,
%\end{equation}
%
where $\mathrm{Precision} = \frac{TP}{TP + FP}$ and $\mathrm{Recall} = \frac{TP}{TP + FN}$.

\noindent \textit{3) LLAMAS.} We use mean average precision (mAP) to evaluate the performance of different algorithms~\cite{llamas2019}.

\noindent
\textbf{Implementation details}.
Since there is no road marking in the upper areas of the input image, we remove the upper part of the original image during both training and testing phases. The size of the processed image is 3384 $\times$ 1010 for ApolloScape, 1640 $\times$ 350 for CULane, and 1276 $\times$ 384 for LLAMAS. 
To save memory usage, we further resize the processed image to 1692 $\times$ 505, 976 $\times$ 208 and 960 $\times$ 288, respectively. We use SGD~\cite{bottou2010large} to train our models and the learning rate is set as 0.01. Batch size is set as 12 for CULane and LLAMAS, and 8 for ApolloScape. The total number of training episodes is set as 80K for CULane and LLAMAS, and 180K for ApolloScape since ApolloScape is more challenging. The cross entropy loss of background pixels is multiplied by 0.4 for CULane and LLAMAS, and 0.05 for ApolloScape since class imbalance is more severe in ApolloScape. For the teacher model, \ie, ResNet-101, we add the pyramid pooling module~\cite{zhao2017pyramid} to obtain both local and global context information. $\alpha_{1}$ and $\alpha_{2}$ are both set as 0.1 and the size of the averaging kernel for obtaining AOI maps is set as 5 $\times$ 5. Our results are not sensitive to the kernel size. 

In our experiments, we use either ERFNet~\cite{romera2017erfnet}, ENet~\cite{paszke2016enet} or ResNet-18~\cite{he2016deep} as the student and ResNet-101 as the teacher. While we choose ERFNet to report most of our ablation studies in this paper, we also report overall results of ENet~\cite{paszke2016enet} and ResNet-18~\cite{he2016deep}. Detailed results are provided in the supplementary material.
We extract both high-level features and middle-level features from ResNet-101 as distillation targets. Specifically, we let the features of block 2 and block 3 of ERFNet to mimic those of block 3 and block 5 of ResNet-101, respectively.
%\cavan{Need more details here, which particular blocks for teacher and student.}

\noindent
\textbf{Baseline distillation algorithms}.
In addition to the state-of-the-art algorithms in each benchmark, we also compare the proposed \algorithmname~with contemporary knowledge distillation algorithms, \ie, KD~\cite{hinton2015distilling}, SKD~\cite{liu2019structured}, PS-N~\cite{yim2017gift}, IRG~\cite{liu2019knowledge} and BiFPN~\cite{zhu2018bidirectional}. Here, KD denotes probability map distillation; SKD employs both probability map distillation and pairwise similarity map distillation; PS-N takes the pairwise similarity map of neighbouring layers as knowledge; IRG uses the instance features, instance relationship and inter-layer transformation of three consecutive frames for distillation, and BiFPN uses attention maps of neighbouring layers as distillation targets.

\begin{table}[!t]
\caption{Performance of different methods on ApolloScape-test.}
\label{apolloscape_table}
\centering
\small{
\begin{tabular}{c|c|c}
\hline
Type & Algorithm & mIoU \\
\hline \hline
\multirow{4}*{Baseline} & Wide ResNet-38~\cite{wu2019wider} & 42.2 \\
~ & ENet~\cite{paszke2016enet} & 39.8 \\
~ & ResNet-50~\cite{he2016deep} & 41.3 \\
~ & UNet-ResNet-34~\cite{ronneberger2015u} & 42.4 \\
\hline
Teacher & \textbf{ResNet-101}~\cite{he2016deep} & \textbf{46.6} \\
\hline
Student & ERFNet~\cite{romera2017erfnet} & 40.4 \\
\hline
\multirow{2}*{Self distillation} & ERFNet-DKS~\cite{sun2019deeply} & 40.8 \\
~ & ERFNet-SAD~\cite{hou2019learning} & 40.9 \\
%\cline{2-3}
\hline
\multirow{6}*{\thead{Teacher-student \\ distillation}} & ERFNet-KD~\cite{hinton2015distilling} & 40.7 \\
~ & ERFNet-SKD~\cite{liu2019structured} & 40.9 \\
~ & ERFNet-PS-N~\cite{yim2017gift} & 40.6 \\
~ & ERFNet-IRG~\cite{liu2019knowledge} & 41.0 \\
~ & ERFNet-BiFPN~\cite{zhu2018bidirectional} & 41.6 \\
\cline{2-3}
%\textbf{R-18-\algorithmname~(ours)} & -- & -- & -- \\
%\hline
%\textbf{R-34-\algorithmname~(ours)} & -- & -- & -- \\
%\hline
~ & \textbf{ERFNet-\algorithmname~(ours)} & \textbf{43.2} \\
\hline
\end{tabular}
}
\vspace{-4ex}
\end{table}

\begin{table}[!t]
\caption{Performance of different methods on CULane-test. To save space, baseline, teacher, student, self distillation and teacher-student distillation in the first column are abbreviated as B, T, S, SD and TSD, respectively.}
\label{culane_table}
\centering
\small{
\begin{tabular}{c|c|c|c|c}
\hline
Type & Algorithm & $F_{1}$ & \thead{Runtime \\ (ms)} & \thead{\# Param \\ (M)} \\
\hline \hline
%Baseline~\cite{llamas2019} & & & & & \\
\multirow{3}*{B} & SCNN~\cite{pan2017spatial} & 71.6 & 133.5 & 20.72 \\
%\cline{2-7}
~ & ResNet-18-SAD~\cite{hou2019learning} & 70.5 & 25.3 & 12.41 \\
~ & ResNet-34-SAD~\cite{hou2019learning} & 70.7 & 50.5 & 22.72 \\
%~ & ResNet-101~\cite{he2016deep} & 71.6 & 171.2 & 52.53 \\
\hline
T & \textbf{ResNet-101}~\cite{he2016deep} & \textbf{72.8} & 171.2 & 52.53 \\
\hline
S & ERFNet~\cite{romera2017erfnet} & 70.2 & \multirow{8}*{\textbf{10.2}} & \multirow{8}*{\textbf{2.49}} \\
\cline{1-3}
\multirow{2}*{SD} & ERFNet-DKS~\cite{sun2019deeply} & 70.6 & ~ & ~ \\
~ & ERFNet-SAD~\cite{hou2019learning} & 71.0 & ~ & ~ \\
\cline{1-3}
%\hline
\multirow{6}*{TSD} & ERFNet-KD~\cite{hinton2015distilling} & 70.5 & ~ & ~ \\
~ & ERFNet-SKD~\cite{liu2019structured} & 70.7 & ~ & ~ \\
~ & ERFNet-PS-N~\cite{yim2017gift} & 70.6 & ~ & ~ \\
~ & ERFNet-IRG~\cite{liu2019knowledge} & 70.7 & ~ & ~ \\
~ & ERFNet-BiFPN~\cite{zhu2018bidirectional} & 71.4 & ~ & ~ \\
\cline{2-3}
~ & \textbf{ERFNet-\algorithmname~(ours)} & \textbf{72.4} & ~ & ~ \\
% Network size & -- & Small & Small & Small & Large & Large & Large \\
% \hline \hline
%Type & ~ & Teacher & Student & Student + distillation & \multicolumn{4}{|c}{Baseline} \\
%\hline
%Category & Proportion & R-101-SAD~\cite{hou2019learning} & ERFNet & \textbf{ERFNet-\algorithmname~} & ENet-SAD~\cite{hou2019learning} & R-18-SAD~\cite{hou2019learning} & R-34-SAD~\cite{hou2019learning} & SCNN~\cite{pan2017spatial} \\
%\hline \hline
%Normal & 27.7\% & \textbf{91.4} & 89.5 & 91.0 & 90.1 & 89.8 & 89.9 & 90.6 \\
%%\hline
%Crowded & 23.4\% & \textbf{71.6} & 67.9 & 71.1 & 68.8 & 68.1 & 68.5 & 69.7 \\
%%\hline
%Night & 20.3\% & \textbf{67.0} & 63.7 & 66.7 & 66.0 & 64.2 & 64.6 & 66.1 \\
%%\hline
%No line & 11.7\% & \textbf{45.5} & 42.1 & 45.0 & 41.6 & 42.5 & 42.2 & 43.4 \\
%%\hline
%Shadow & 2.7\% & \textbf{71.0} & 66.2 & 70.8 & 65.9 & 67.5 & 67.7 & 66.9 \\
%%\hline
%Arrow & 2.6\% & \textbf{88.4} & 83.6 & 86.3 & 84.0 & 83.9 & 83.8 & 84.1 \\
%%\hline
%Dazzle light & 1.4\% & \textbf{66.9} & 59.9 & 65.0 & 60.2 & 59.8 & 59.9 & 58.5 \\
%%\hline
%Curve & 1.2\% & \textbf{66.3} & 65.7 & 65.7 & 65.7 & 65.5 & 66.0 & 64.4 \\
%%\hline
%Crossroad & 9.0\% & 2732 & 2056 & 2542 & 1998 & 1995 & \textbf{1960} & 1990 \\
%%\hline
%Total & -- & \textbf{72.8} & 70.2 & 72.4 & 70.8 & 70.5 & 70.7 & 71.6 \\
%\hline \hline
%Runtime (ms) & -- & 171.2 & \textbf{10.2} & \textbf{10.2} & 13.4 & 25.3 & 50.5 & 133.5  \\
%%\hline
%Parameter (M) & -- & 52.53 & 2.49 & 2.49 & \textbf{0.98} & 12.41 & 22.72 & 20.72  \\
\hline
\end{tabular}
}
\vspace{-2ex}
\end{table}

\begin{table}[!t]
\caption{Performance of different methods on LLAMAS-test.}
\label{llamas_table}
\centering
\small{
\begin{tabular}{c|c|c}
\hline
Type & Algorithm & mAP \\ %& AP$_{L1}$ & AP$_{L0}$ & AP$_{R0}$ & AP$_{R1}$ \\
\hline \hline
%Baseline~\cite{llamas2019} & & & & & \\
\multirow{3}*{Baseline} & SCNN~\cite{pan2017spatial} & 0.597 \\ %& 0.321 & 0.874 & 0.866 & 0.326 \\
%\cline{2-7}
~ & ResNet-50~\cite{he2016deep} & 0.578 \\ %& 0.308 & 0.838 & 0.842 & 0.322 \\
~ & UNet-ResNet-34~\cite{ronneberger2015u} & 0.592 \\ %& 0.310 & 0.842 & 0.854 & 0.363 \\
\hline
Teacher & \textbf{ResNet-101}~\cite{he2016deep} & \textbf{0.607} \\ %& \textbf{0.315} & \textbf{0.876} & \textbf{0.864} & \textbf{0.371} \\
\hline
Student & ERFNet~\cite{romera2017erfnet} & 0.570 \\ %& 0.303 & 0.836 & 0.827 & 0.314 \\
\hline
\multirow{2}*{Self distillation} & ERFNet-DKS~\cite{sun2019deeply} & 0.573 \\ %& 0.305 & 0.840 & 0.828 & 0.319 \\
~ & ERFNet-SAD~\cite{hou2019learning} & 0.575 \\ %& 0.307 & 0.842 & 0.830 & 0.321 \\
%\cline{2-7}
\hline
\multirow{6}*{\thead{Teacher-student \\ distillation}} & ERFNet-KD~\cite{hinton2015distilling} & 0.572 \\ %& 0.305 & 0.839 & 0.827 & 0.318 \\
~ & ERFNet-SKD~\cite{liu2019structured} & 0.576 \\ %& 0.310 & 0.844 & 0.830 & 0.321 \\
~ & ERFNet-PS-N~\cite{yim2017gift} & 0.575 \\ %& 0.306 & 0.845 & 0.831 & 0.318 \\
~ & ERFNet-IRG~\cite{liu2019knowledge} & 0.576 \\ %& 0.308 & 0.843 & 0.832 & 0.320 \\
~ & ERFNet-BiFPN~\cite{zhu2018bidirectional} & 0.583 \\ %& 0.314 & 0.853 & 0.839 & 0.327 \\
\cline{2-3}
~ & \textbf{ERFNet-\algorithmname~(ours)} & \textbf{0.598} \\ %& \textbf{0.315} & \textbf{0.874} & \textbf{0.856} & \textbf{0.361} \\
\hline
\end{tabular}
}
\vspace{-4ex}
\end{table}

\subsection{Results}

Tables~\ref{apolloscape_table}-~\ref{llamas_table} summarize the performance of our method, \ie, ERFNet-\algorithmname~, against state-of-the-art algorithms on the testing set of ApolloScape~\cite{huang2018apolloscape}, CULane~\cite{pan2017spatial} and LLAMAS~\cite{llamas2019}. We also report the runtime and parameter size of different models in Table~\ref{culane_table}. The runtime is recorded using a single GPU (GeForce GTX TITAN X Maxwell)
and averages across 100 samples. 
ERFNet-\algorithmname~outperforms all baselines and previous distillation methods in all three benchmarks. Note that ERFNet-\algorithmname~has 21 $\times$ fewer parameters and runs 16 $\times$ faster compared with ResNet-101 on CULane testing set; the appealing performance strongly suggests the effectiveness of \algorithmname.

\begin{figure}[t]
  \centering
  \includegraphics[width=1.0\linewidth]{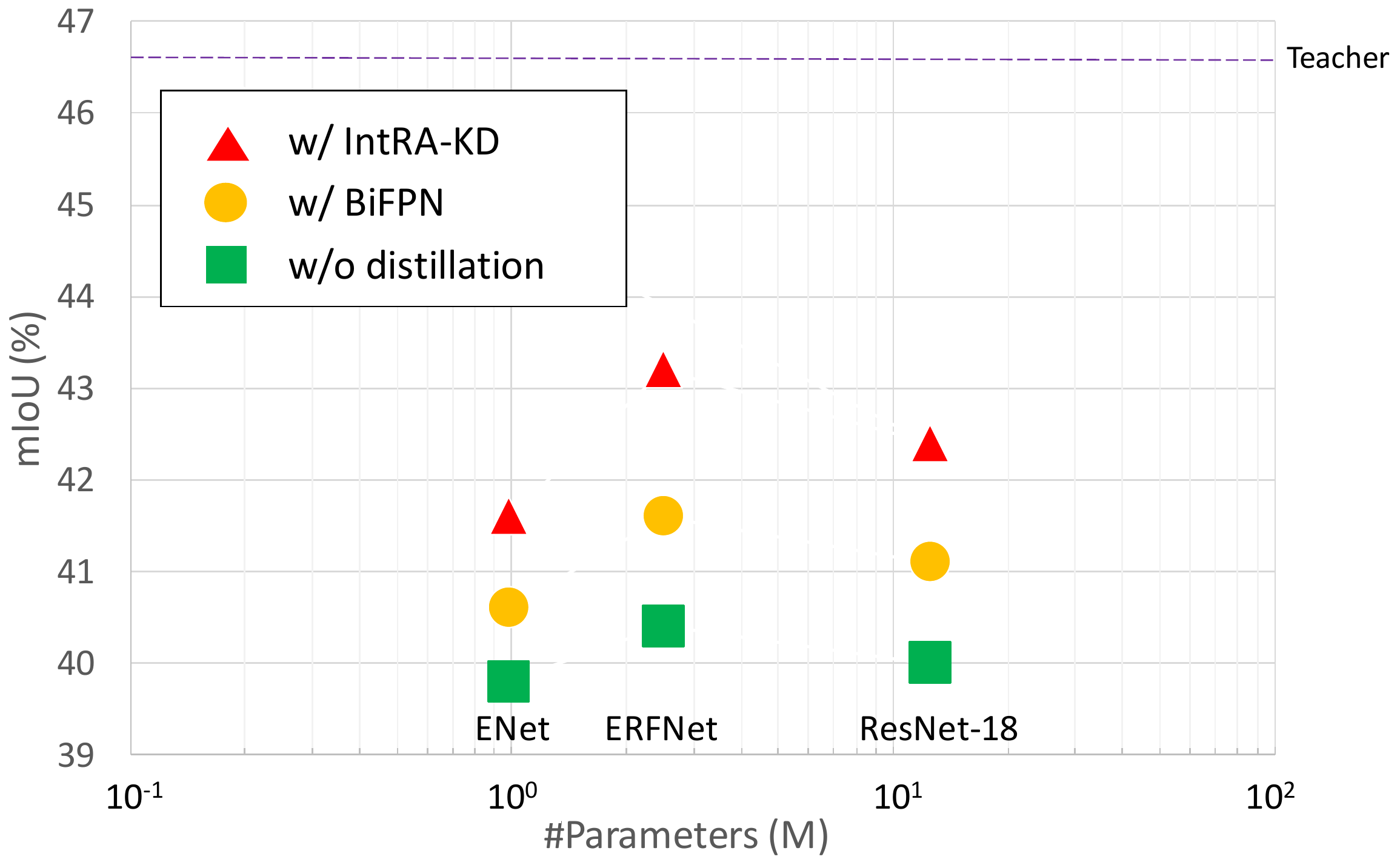}
  \vskip -0.2cm
  \caption{Comparison between \algorithmname~and BiFPN on ENet, ERFNet and ResNet-18 on ApolloScape-test.}
  \centering
  \vskip -0.2cm
  \label{fig:log}
\end{figure}

We also apply \algorithmname~to ENet and ResNet-18, and find that \algorithmname~can equivalently bring more performance gains to the backbone models than the state-of-the-art BiFPN~\cite{zhu2018bidirectional} on ApolloScape dataset (Fig.~\ref{fig:log}). Note that BiFPN is a competitive algorithm in all benchmarks. Encouraging results are also observed on CULane and LLAMAS when applying \algorithmname~to ENet and ResNet-18. Due to space limit, we report detailed performance of applying different distillation algorithms to ENet and ResNet-18 in the supplementary material. The effectiveness of our \algorithmname~on different backbone models validates the good generalization ability of our method.   
%Moreover, \algorithmname~increases the $F_{1}$-measure of ENet and ResNet-18 from 69.8, 69.4 to 71.8 and 71.4 on CULane testing set. \algorithmname~also increases the mAP of ENet and ResNet-18 from 0.562, 0.565 to 0.585 and 0.590 on LLAMAS testing set. The effectiveness of our \algorithmname~on different backbone models validates the good generalization ability of our method. Due to space limit, we report detailed performance of applying different distillation algorithms to ENet and ResNet-18 in the supplementary material. 

\begin{figure}[t]
  \centering
  \includegraphics[width=\linewidth]{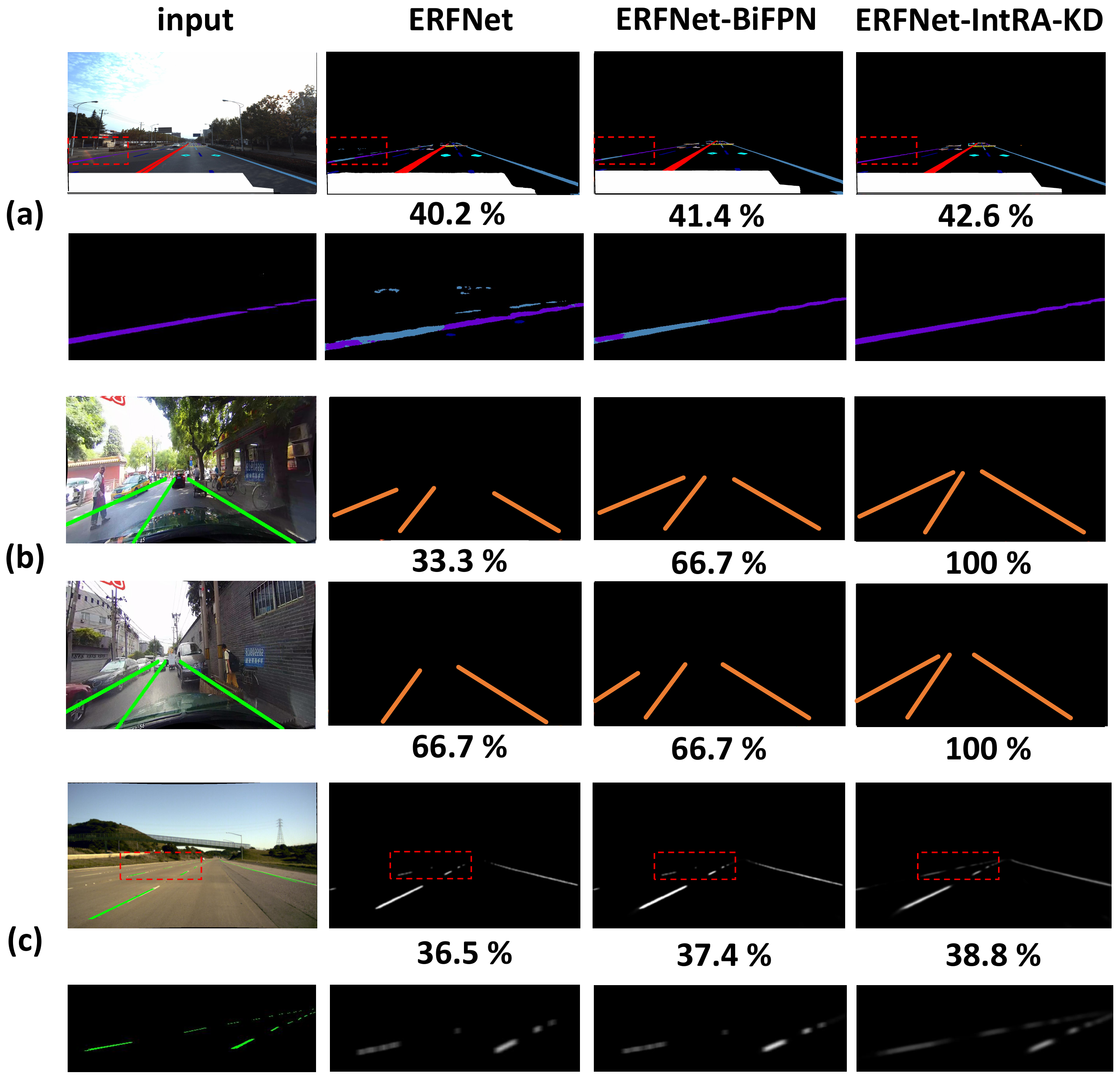}
  \vskip -0.1cm
  \caption{Performance of different methods on (a) ApolloScape, (b) CULane and (c) LLAMAS testing sets. The number below each image denotes the accuracy for (a) and (c), $F_{1}$-measure for (b). Ground-truth labels are drawn on the input image. Second rows of (a) and (c) are enlarged areas covered by the red dashed rectangle.}
  \centering
  \vskip -0.6cm
  \label{fig:qualitative_result}
\end{figure}

\begin{figure*}[t]
  \centering
  \includegraphics[width=1.0\linewidth]{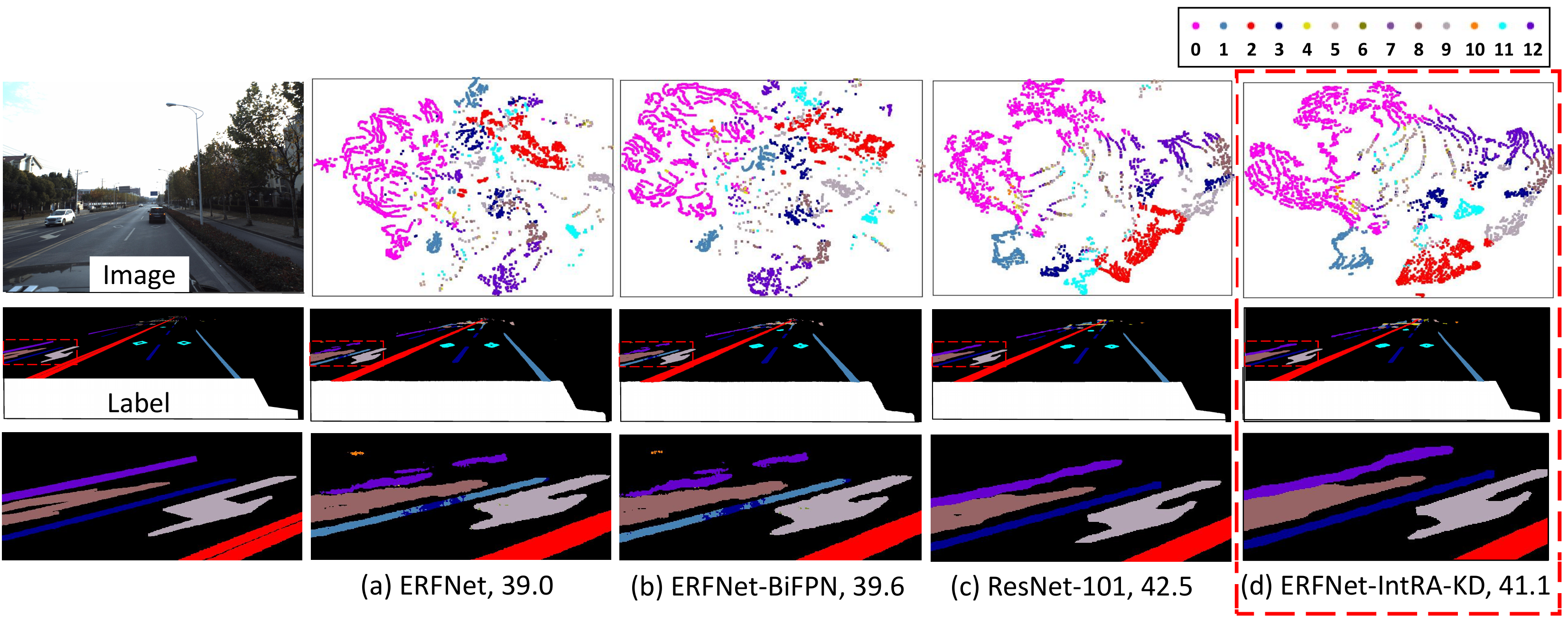}
  \vskip -0.2cm
  \caption{Deep feature embeddings (first row) and predictions (second row) of (a) ERFNet (b) ERFNet-BiFPN (c) ResNet-101 (teacher) (d) ERFNet-\algorithmname. The number next to the model's name denotes accuracy ($\%$). Regions of the model prediction, which are covered by the red dashed rectangle, are highlighted in the third row. The color bar of the deep embeddings is the same as that of the model prediction except the background, whose color is changed from black to pink for better visualization. Note that we crop the upper part of the label and model prediction for better visualization and we use t-SNE to visualize the feature maps (first row).}
  \centering
  \vskip -0.4cm
  \label{fig:tsne_map}
\end{figure*}

We also show some qualitative results of our \algorithmname~and BiFPN~\cite{zhu2018bidirectional} (the most competitive baseline) on three benchmarks. As shown in (a) and (c) of Fig.~\ref{fig:qualitative_result}, \algorithmname~helps ERFNet predict more accurately on both long and thin road markings. As to other challenging scenarios of crowded roads and poor light conditions, ERFNet and ERFNet-BiFPN either predict lanes inaccurately or miss the predictions. By contrast, predictions yielded by ERFNet-\algorithmname~are more complete and accurate.

Apart from model predictions, we also show the deep feature embeddings of different methods. As can be seen from Fig.~\ref{fig:tsne_map}, the embedding of ERFNet-\algorithmname~is more structured compared with that of ERFNet and ERFNet-BiFPN.
In particular, the features of ERFNet-\algorithmname~are more distinctly clustered according to their classes in the embedding, with similar distribution to the embedding of the ResNet-101 teacher.
The results suggest the importance of structural information in knowledge distillation.

%the prediction of ERFNet model is messy especially on these small road markings while the prediction of the large teacher model is much clearer. \algorithmname~can make ERFNet predict more accurately on small road markings than BiFPN. Besides, \algorithmname~makes the deep feature embeddings of ERFNet more distinctly clustered according to their classes. The above results strongly indicate that the structural information conveyed by \algorithmname~is vital to improving the predictions of the ERFNet model.

\subsection{Ablation Study}

In this section, we investigate the effect of each component, \ie, different loss terms and the associated coefficients, on the final performance.
%Besides, we also combine AOI with previous distillation methods and find that it can bring consistent performance gains. 

\begin{figure}[t]
  \centering
  \includegraphics[width=1.0\linewidth]{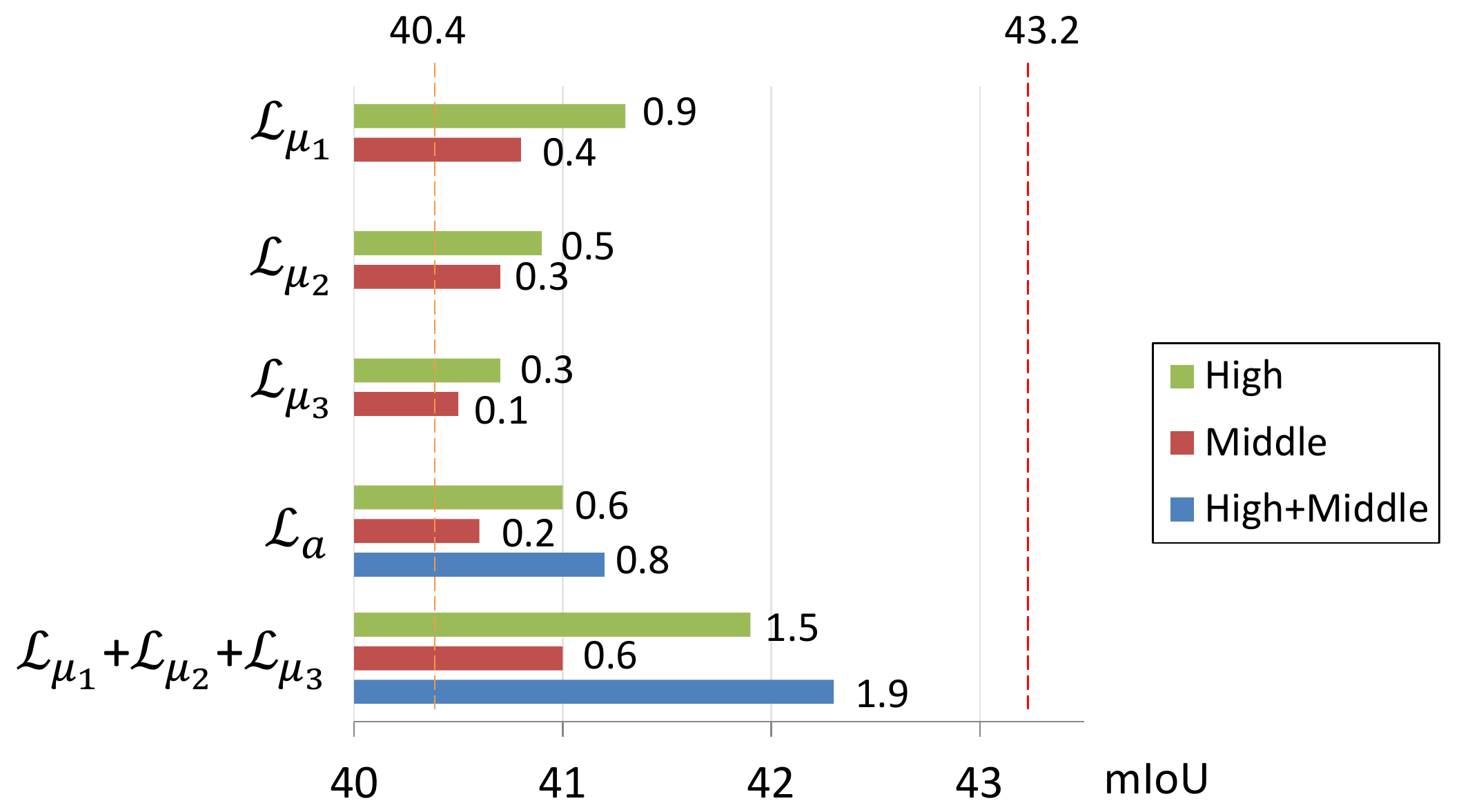}
  %\vskip -0.2cm
  \caption{Performance (mIoU) of ERFNet using different loss terms of \algorithmname~on ApolloScape-test. Here, ``High" denotes the mimicking of high-level features (block 5) and ``Middle" denotes the mimicking of middle-level features (block 3) of the teacher model. $\mathcal{L}_\mathrm{\mu_{i}}$ denotes the variant where only the $i$-th order moment is deployed for inter-region affinity distillation. Here, ``40.4" is the performance of ERFNet without distillation and ``43.2" is the performance of ERFNet-\algorithmname~ that considers $\mathcal{L}_{\mu_1}+\mathcal{L}_{\mu_2}+\mathcal{L}_{\mu_3}$ and $\mathcal{L}_{a}$. The number besides each bar is performance gain brought by each loss term compared with ERFNet.}
  \centering
  \vskip -0.7cm
  \label{fig:loss}
\end{figure}

\noindent \textbf{Effect of different loss terms.} From Fig.~\ref{fig:loss}, we have the following observations: (1) Considering all moments from both middle- and high-level features, \ie, the blue bar with $\mathcal{L}_{\mu_1}+\mathcal{L}_{\mu_2}+\mathcal{L}_{\mu_3}$, brings the most performance gains. (2) Attention map distillation, $\mathcal{L}_{a}$ also brings considerable gains compared with the baseline without distillation. (3) Distillation of high-level features brings more performance gains than that of middle-level features. This may be caused by the fact that high-level features contain more semantic-related information, which is beneficial to the segmentation task. (4) Inter-region affinity distillation and attention map distillation are complementary, leading to best performance (\ie, 43.2 mIoU as shown by the red vertical dash line).

%\begin{table}[!t]
%\caption{Performance of ERFNet using different loss terms of \algorithmname~on ApolloScape testing set.}
%\label{ablation_table}
%\centering
%\small{
%\begin{tabular}{c|c}
%\hline
%Loss term & mIoU (relative gain) \\
%\hline \hline
%None & 40.4 (0.0) \\
%%\hline
%%+ $\mathcal{L}_\mathrm{feature}$ & 41.2 (0.8) \\
%\hline
%+ $\mathcal{L}_\mathrm{a}$ & 41.4 (0.8) \\
%\hline
%+ $\mathcal{L}_\mathrm{m}$ & 42.3 (1.9) \\
%\hline
%+ $\mathcal{L}_\mathrm{a}$ + $\mathcal{L}_\mathrm{m}$ & 43.2 (2.8) \\
%\hline
%\end{tabular}
%}
%\vspace{-2ex}
%\end{table}

\noindent \textbf{Effect of loss coefficients.} The coefficients of the attention map loss and affinity distillation loss are all set as 0.1 to normalize the loss values. Here, we test different selections of the loss coefficients, \ie, selecting coefficient value from $\{0.05, 0.10, 0.15\}$. ERFNet-\algorithmname~achieves similar performance in all benchmarks, \ie, $\{43.18, 43.20, 43.17\}$ mIoU in ApolloScape, $\{72.36, 72.39, 72.38\}$ $F_{1}$-measure in CULane and $\{0.597, 0.598, 0.598\}$ mAP in LLAMAS. Hence, \algorithmname~is robust to the loss coefficients.

%\noindent \textbf{Combination of AOI with previous distillation methods.} From Table~\ref{aoi_distillation_table}, it is evident that AOI can consistently bring considerable performance gains to all previous distillation methods. This is apparent since the distillation process is focused around road markings and the neighbouring areas.
%
%\begin{table}[!t]
%\caption{Performance of combining different knowledge distillation methods with AOI on ApolloScape testing set.}
%\label{aoi_distillation_table}
%\centering
%\small{
%\begin{tabular}{c|c|c}
%\hline
%Backbone & Algorithm & mIoU (relative gain) \\
%\hline \hline
%\multirow{11}*{ERFNet~\cite{romera2017erfnet}} & None & 40.4 (0.0)\\ 
%\cline{2-3}
%~ & KD~\cite{hinton2015distilling} & 40.7 (0.3) \\
%\cline{2-3}
%~ & KD + AOI & 41.1 (0.7) \\
%\cline{2-3}
%~ & SKD~\cite{liu2019structured} & 40.9 (0.5) \\
%\cline{2-3}
%~ & SKD + AOI & 41.6 (1.2) \\
%\cline{2-3}
%~ & PS-N~\cite{yim2017gift} & 40.6 (0.2) \\
%\cline{2-3}
%~ & PS-N + AOI & 41.1 (0.7) \\
%\cline{2-3}
%~ & IRG~\cite{liu2019knowledge} & 41.0 (0.6) \\
%\cline{2-3}
%~ & IRG + AOI & 41.6 (1.2) \\
%\cline{2-3}
%~ & BiFPN~\cite{zhu2018bidirectional} & 41.6 (1.2) \\
%\cline{2-3}
%~ & BiFPN + AOI & 42.2 (1.8) \\
%\hline
%\end{tabular}
%}
%\vspace{-2ex}
%\end{table}